\documentclass{article}
\usepackage{enumerate}
\usepackage{multirow}
\usepackage[bottom,flushmargin,hang,multiple]{footmisc}
\usepackage{etoolbox}
\makeatletter
\patchcmd\maketitle{\hb@xt@1.8em}{\hbox}{}{}
\makeatother
\patchcmd{\maketitle}{\@fnsymbol}{\@arabic}{}{}
\usepackage{amsmath}
\usepackage{graphicx}
\usepackage{caption}
\usepackage{subcaption}
\usepackage[margin=1.5in]{geometry}
\usepackage{float}
\usepackage{comment}
\usepackage{pgf}
\usepackage{tikz}
\usepackage{subcaption}

\usepackage[T1]{fontenc}
\usepackage[sfdefault,scaled=.95,light]{FiraSans}
\usepackage{newtxsf}

\usepackage{etoolbox}
\AtBeginEnvironment{quote}{\small}

\begin{document}

\title{BEST : A decision tree algorithm that handles missing values}



\author{C\'edric Beaulac \footnote{ University of Toronto \newline Department of Statistical Sciences  \newline E-mail : beaulac.cedric@gmail.com } \and Jeffrey S. Rosenthal \footnote{University of Toronto \newline Department of Statistical Sciences  \newline E-mail : jeff@math.toronto.edu }  
}


\date{\today}

\maketitle

\begin{abstract}
The main contribution of this paper is the development of a new decision tree algorithm. The proposed approach allows users to guide the algorithm through the data partitioning process. We believe this feature has many applications but in this paper we demonstrate how to utilize this algorithm to analyse data sets containing missing values. We tested our algorithm against simulated data sets with various missing data structures and a real data set. The results demonstrate that this new classification procedure efficiently handles missing values and produces results that are slightly more accurate and more interpretable than most common procedures without any imputations or pre-processing.

\bigskip


\end{abstract}

\pagebreak

\section{Introduction}

Machine learning algorithms are used in many exciting real data applications, but may have problems handling predictors with missing values. Imputation techniques were designed to handle data with missing value under the missing completely at random (MCAR) assumption. Since this is a restrictive assumption we propose a solution to missing values that uses the tree structure of Classification and Regression Trees (CART) to deal in an intuitive manner with observations that are missing in patterns which are not completely at random.

\bigskip

Our proposed new tree construction procedure was inspired by a data set where the missing pattern of one subset of predictors could be perfectly explained by another subset (see Section \ref{ME}). A typical decision tree is an algorithm that partitions the predictor space based upon a predictor value, splitting it into two subspaces and repeats this process recursively. 

\bigskip

Our proposed algorithm is different as it allows the researcher to impose a structure on the variables available for the partitioning process. By doing so, we construct Branch-Exclusive Splits Trees (BEST). When a predictor $X_j$ contains missing values, we can use other predictors to identify the region where the predictor $X_j$ contains no missing value. Therefore we can use the proposed algorithm to consider splitting on a predictor only when it contains no missing value based on previous partitioning. BEST can be easily adapted to any splitting rule and any forest forming procedure \cite{Breiman96,Breiman01,Geurts06}. BEST also has other applications; it can be used by researchers that would like to utilize some knowledge they have on the data generating distribution in order to guide the algorithm in selecting a more accurate and more interpretable classifier. 

\bigskip

In this article we will briefly discuss the classification problem and its notation and we will explain how classification trees solve that problem. Afterwards, we will introduce the proposed algorithm and some motivating examples before explaining in detail how the algorithm functions. We will then do a quick review of the literature to position our algorithm within the current literature. Finally, some tests will be performed on simulated data sets and on the real data that inspired this new algorithm.

\section{The classification problem} \label{CP}

In a typical supervised statistical learning problem we are interested in understanding the relationship between a response variable $\mathbf{Y}$ and an associated $m$-dimensional predictor vector $\mathbf{X} = (X_1,...,X_m)$. When the response variable is categorical and takes $k$ different possible values, this problem is defined as a $k$-class classification problem. In that set up, an interesting challenge is to use a data set $S = \{ (y_i,x_{i,1},...,x_{i,m}) ; i = 1,...,n \}$ in order to construct a classifier $h$. Most of the time, it is assumed that the observations within our data set were drawn independently from the same unknown and true distribution $\mathcal{D}$, i.e. $\mathbf{X} \times \mathbf{Y} \sim \mathcal{D}$. A classifier is built to emit a class prediction for any new data point $X$ that belongs in the predictor space $\mathcal{X} = \mathcal{X}_1 \times ... \times \mathcal{X}_m$. Therefore a classifier divides the predictor space $\mathcal{X}$ into $k$ disjoint regions $R_1,...R_k$, one per class, such that $\cup_{q =1}^k R_q = \mathcal{X}$, i.e. $h(x) = \sum_{q=1}^k q \mathbf{1}\{ x \in R_q\}$.

\bigskip

\subsection{Classification and Regression Trees} \label{CT}

A classification tree \cite{Breiman84} is an algorithm that forms regions in the predictor space by recursively dividing it, more precisely, this procedure performs recursive binary partitioning. Beginning with the entire predictor space, the algorithm selects the variable to split upon and the location of the split that minimize some impurity measure. Then the resulting two regions are each split into two more regions until some stopping rule is applied. The classifier will label each region with one of the $k$ possible classes.

\bigskip

The traditional labelling process goes as follows; let $p_{rq} = \frac{1}{n_r} \sum_{x_i \in R_r} \mathbf{1}\{y_i = q\}$, the proportion of the class $q$ in the region $r$ where $n_r$ is the number of observations contained in region $r$. Then, the label of the region $r$ is the majority class in that region, i.e. if $x \in R_r$, $h_S(x) = \textrm{argmax}_q (p_{rq})$. For regression trees, the output mean within a leaf region is used as prediction for observations that belong in that region. The impurity measure function for region $r$ is defined as $Q_r$ and can take many forms such as the \textit{Gini index}, the \textit{deviance} or the \textit{misclassification error}. For regression trees, the \textit{mean squared error} is one possible region impurity measure.

\bigskip

When splitting a region $R_p$ into two new regions $R_r$ and $R_t$ the algorithm will compute the total impurity of the new regions ; $ n_{r} Q_r + n_t Q_t$ and will pick the split variable $X_j$ and split location $s$ that minimizes that total impurity. If the predictor $X_j$ is continuous, the possible splits are of the form $X_{j} \leq s$ and $X_j > s$ which usually results in $n_p-1$ possible splits. For a categorical predictor having $c$ possible values, we usually consider all of the $2^{c-1} -1$ possible partitions. 

\bigskip

The partitioning continues until a stopping rule is applied. In some cases, the algorithm stops whenever every terminal node of the tree contains less than $\beta$ observations, in other cases it stops when all observations within a region belong to the same class. To prevent overfitting, a deep tree is built and then the tree can be pruned. Tree-pruning is a cost-complexity procedure that relies on considering that each leaf, region, is associated with a cost $\alpha$. The procedure begins by collapsing leaves that produce the smallest increase in total impurity and this technique will collapse leaves as long as the increase in impurity is less than the cost $\alpha$ of the additional leaf. The $\alpha$ parameter can be determined by cross-validation or with the use of a validation set. 

\setlength\parindent{15pt}

\section{Missing values} \label{NA}

Let us now introduce the definition of missing data we are using in this article. As described in the previous section, a standard assumption in data analysis is that all observations are distributed according to the true data generation distribution $\mathcal{D}$. We could think of the missingness itself as a random variable $\mathbf{M}$ also of dimension $m$ that is distributed according to some missingness generating distribution which is a part of $\mathcal{D}$, i.e. $\mathbf{X} \times \mathbf{M} \times \mathbf{Y} \sim \mathcal{D}$. Formally, if $\mathbf{M}$ represents the missingness of the vector of predictors $\mathbf{X}$ it means that $M_j = 1$ if $X_j$ is observed and $M_j = 0$ if $X_j$ is missing.

\bigskip

Three different relationships between $\mathbf{M}$ and $\mathbf{X}$ were defined by Rubin \cite{Rubin76} and by Little and Rubin \cite{Rubin02}. Seaman \cite{Seaman13} later untangled the many definition inconsistencies of these relationships. In this article, we will rely on simple definitions for an easy understanding of the structure we will consider. First, missing completely at random (MCAR) is the simplest structure we consider:  $\mathbf{M} \perp \mathbf{X}$. Here, we consider the data is MCAR if the set of missing patterns $M$ is independent of the set of predictors. 

\bigskip

Missing at random (MAR) is much more complicated; it essentially means that the missingness $\mathbf{M}$ is independent of missing observations but can still depend on observed predictors. More rigorously, we define $\text{X}^{o} = \{ x_{ij} \in S | x_{ij} \text{ is observed} \}$ as the set of all observed predictors value, and $\text{X}^{na} = \{ x_{ij} \in S | x_{ij} \text{ is missing} \}$ as the set of missing predictors value. We say that data is MAR if the distribution of the missingness is conditionally independent of missing values given observed values : $\mathbf{M} \perp \text{X}^{na} |\text{X}^{o}$. As pointed by Seaman \cite{Seaman13}, MAR has not always been used consistently and the definition above is the one we settled on for this project. Note that MCAR implies MAR. 

\bigskip

Finally, if the missing data is neither MCAR nor MAR, we say that the data is missing not at random (MNAR). We will see that the relationship between $\mathbf{M}$ and $\mathbf{X}$ has a considerable effect on the efficiency of the missing values techniques that exist. In the sections to come we will test our algorithms under these various missing data structures.

\section{Branch-Exclusive Splits Trees (BEST)} \label{BEST}

We now introduce the proposed algorithm, BEST. The purpose of BEST is to utilize the tree structure itself in order to manage some missing data or some special structure among predictors. 

\subsection{Motivating Example} \label{ME}

To begin we will explain which data structure BEST is suitable for by introducing the motivating data set. It contains information regarding the academic performances of students. The data set was provided to us by the Univeristy of Toronto and was first introduced and analysed by Bailey et al. \cite{Bailey16}. It was later analysed by Beaulac and Rosenthal \cite{Beaulac18} where the goal was to predict whether or not a student would complete its program. The predictors represent the number of credits and grades obtained in all the departments during the first two semesters. Understanding the importance of these predictors was also a question raised by the authors. Table \ref{datatable} contains a preview of the data with a reduced number of departments.

\begin{table}[H] 
\begin{center}
\resizebox{\textwidth}{!}{
\begin{tabular}{ |c|c|c|c|c|c|c| } 
\hline
 Student ID  & Credits Math & Grade Math & Credits Econ &  Grade  Econ & Credits Hist & Grade  Hist  \\
\hline
101  & 2 & 72 &  3 & 88 & 0 & NA \\
  \hline
208  & 0 & NA &  0 & NA & 5 & 78 \\
\hline
\end{tabular}}
\end{center}
\caption{An example of the motivating data set for 2 students and 3 departments} 
\label{datatable} 
\end{table}

\begin{figure}[H]
\centering
\begin{tikzpicture}[
    scale = 1, transform shape, thick,
    every node/.style = {draw, rectangle, rounded corners, minimum size = 10mm},
    grow = down,  
    level 1/.style = {sibling distance=6cm},
    level 2/.style = {sibling distance=4cm}, 
    level 3/.style = {sibling distance=2cm}, 
    level 4/.style = {sibling distance=2cm}, 
    level distance = 2cm
  ]
  \node (Start){Grade not available} {
   child  { node (A) {Grade not available}  
   child { node(B){Grade not available}}
   child {node(C){Grade available}}
   }
   child {   node  (D) {Grade available}   }
  };

  \begin{scope}[nodes = {draw = none}]
    \path (Start) -- (A) node [near start, left]  { Credit $< 2$};
    \path (A) -- (B) node [near start, left]  {Credit $= 0$};
    \path (A) -- (C) node [near start, right]  { Credit $> 0$};
    \path (Start) -- (D) node [near start, right] { Credit $\geq 2$};

  \end{scope}
\end{tikzpicture}
\caption{An example of BEST decision tree partitioning upon the Credit variable and the availability of the Grade variable within each of the produced regions \label{GradeTree}}
\end{figure}
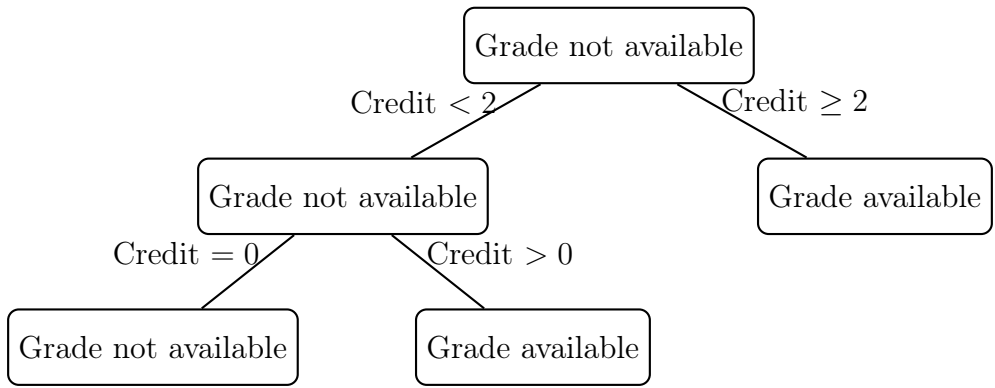

A student has no grade in many departments as they can only register to a limited number of courses within a year. In this situation, many grade variables are missing for every student. BEST handles that problem by considering the averaged grade obtained in a department only for students who took courses in that department. For example, BEST will force the classification tree algorithm to split upon the \textit{Number of credits} predictor to begin. Then, suppose \textit{Number of credits in Statistics} is selected and 2 is the split point for the partitioning, BEST will then allow splits on the \textit{Grade in Statistics} predictor for the group of students in the region defined by \textit{Number of credits in Statistics} $> 2$. Therefore, the \textit{Number of credits} variables are used to define the region where the respective \textit{Grade} variables are available for the partitioning process and thus we say that the \textit{Number of credits} are gating variables for the \textit{Grade} variables. The many \textit{Grade} variables are gated by their respective \textit{Number of credits} variables. Figure \ref{GradeTree} illustrates how a BEST decision tree uses  \textit{Number of credits} as a gating variable for \textit{Grade}. 

\bigskip

\bigskip  

Other real data sets with similar problems are surveys. Many surveys have questions that are only relevant based on previous answers. Suppose question \#1 is yes/no and is followed by : \textit{If you answered no, please go to question \#3}. This is quite typical and in that situation, BEST can use the yes/no question as a gating variable for the question \#2.

\subsection{Intuition}

As we explained in section \ref{CT}, a classification tree aims at partitioning the predictor space and labelling the resulting regions. CART does so by looking through all the possible splits and selecting the one that minimizes some pre-specified error measure. When using BEST some predictors are available to split upon only within some regions of the predictor space, such as the \textit{Grade} variables in the motivating data set. These regions are defined according to other predictors, such as the \textit{Number of credits} variables in the motivating data set. More generally, predictor $X_l$ could be only available for the partitioning process in the region defined by $X_j < t$. We say that $X_j$ is a gating variable for $X_l$ or that $X_l$ is gated by $X_j$. The variable $X_l$ will not be available for the partitioning process until the gating variable allows it. Table \ref{splittable} illustrates when BEST can partition the data using $X_l$ based on a previous partitioning where BEST selects $X_j$ as the splitting variable and $s$ as the splitting value.

\bigskip
 
\begin{table}[H] 
\begin{center}
\resizebox{\textwidth}{!}{
\begin{tabular}{ |c|c|c|c|c| } 
\hline
   & \multicolumn{2}{|c|}{$s<t$}  & \multicolumn{2}{|c|}{$s > t$}  \\
\hline
  & Region $X_j < s$ & Region $X_j \geq s$ &  Region $X_j < s$ & Region $X_j \geq s$ \\
  \hline
 $X_l$ is available & Yes & No & No & No \\
\hline
\end{tabular}}
\end{center}
\caption{Availability of $X_l$ if $X_j$ is previously selected as splitting variable and $s$ as splitting value} 
\label{splittable} 
\end{table}

\bigskip

\begin{figure}[H]
\centering
\begin{tikzpicture}[
    scale = 1, transform shape, thick,
    every node/.style = {draw, rectangle, rounded corners, minimum size = 10mm},
    grow = down,  
    level 1/.style = {sibling distance=5cm},
    level 2/.style = {sibling distance=2cm}, 
    level 3/.style = {sibling distance=2cm}, 
    level 4/.style = {sibling distance=2cm}, 
    level distance = 2cm
  ]
  \node (Start){$X_j$ not available} {
   child  { node (A) {$X_j$ not available}   }
   child {   node  (D) {$X_j$ available}   }
  };

  \begin{scope}[nodes = {draw = none}]
    \path (Start) -- (A) node [near start, left]  {$M_j = 0$};
    \path (Start) -- (D) node [near start, right] {$M_j = 1$};
  \end{scope}
\end{tikzpicture}
\caption{An example of BEST decision tree partitioning upon the dummy variable $M_j$ that defines the availability of $X_j$ \label{NoGating}}
\end{figure}
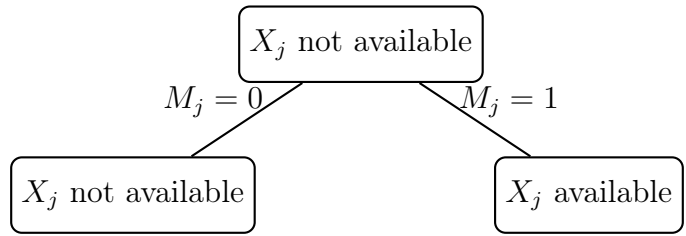

Doing so, predictors with missing values can be handled easily as BEST will partition the data according to that predictor only in regions where it does not contain missing value. If a data set contains missing values on predictor $X_j$ but no predictor can help define the region with no missing value, we can add a new predictor $X_{m+1}$ to the model as our gating variable. This new predictor is a dummy variable such that $X_{i,m+1} = 0$ if  $X_{i,j}$ is missing and 1 if not. Doing so, we effectively add $M_j$ as defined in section \ref{NA}, as a predictor in the model and thus will be defined as follows in the rest of the text. Then, BEST will only consider splitting on $X_j$ in the subspace defined by $M_{j} = 1$ as illustrated in figure \ref{NoGating}. Multiple dummy variables are added to the model if multiple predictors contain missing values. Doing so also allows us to analyse the individual importance of the missing patterns $M$.

\bigskip

Finally, some insight on the data structure can be used to force some variable to be partitioned upon before others which is another application of BEST not described in this article. The result is a tree-structured classification model where some split variables are branch-exclusives. Even though we do not further mention it, the construction described below could be used for regression trees as well.

\subsection{Algorithm implementation}

Let us now explain how the algorithm functions. BEST takes as input the full data set $S$, the tuning parameter $\beta$ and a list containing the predictor availability structure $V$. First $S$ is set as the root node, the first set of observations to go through the following steps. The algorithm verifies a set of conditions before proceeding with the partitioning process. The first condition (C1) is that the region contains more than $\beta$ observations, this is the main stopping rule. Then, the next condition (C2) is that the observations in the region have different labels; this condition makes sure that the algorithm has a reason to partition the data. Finally, the last condition (C3) is that at least one of the available predictors takes different values among the observations in the region; this is to guarantee that the algorithm can actually partition the data. 

\bigskip

If at least one condition is false, then the region is defined as a leaf, a label is assigned to that leaf for prediction purposes and the partitioning process is stopped. Usually the class that represents the majority in a leaf is selected as the label for that region, but one could define different label assignment rules.

\bigskip

If all conditions (C1, C2 and C3) are respected then the partitioning process begins. The algorithm will go through all available predictors. For a predictor $j$, the algorithm will go through all possible partitions $s$ of the region with respect to the predictor $j$ and will compute the total impurity of the resulting two regions $ n_{r_1} Q_{r_1} + n_{r_2} Q_{r_2}$. Any region impurity measure $Q$ can be used. BEST then selects the predictor $j$ and the split $s$ that minimize the total impurity and creates two children regions by splitting the data according to $s$. 

\bigskip

The last step is to update the list of available predictors for the children regions. There exist multiple possible structures that can contain this information but within the R-language \cite{R} we have settled on a list $V$ since lists are very flexible, but other structure could have been used. To begin $V[1]$ represents the set of predictors available for the partitioning process in the root node. More specifically $V[1]$ is a vector of size $m$ where $V[1][j] = 1$ if the $j$th predictor is available to be split upon in the root node and $V[1][j] = 0$ otherwise. For instance, for the data set introduced in section \ref{ME}, the vector $V[1]$ equal 0 for the \textit{Grade} variables since they are not available for the partitioning process initially.

\bigskip

In the meanwhile, further elements of the list such as $V[j+1]$ are defined for $j \in 1,...m$ and they contain the necessary information to update the predictors available for further partitioning. If $j$ is a gating variable, then $V[j+1]$ should reflect that and contain the information needed to update the variable available following a partitioning based on $j$. For instance, if $j$ is a continuous predictor, $V[j+1]$ contains a threshold value and a list this of variables made available from the appropriate partitioning. For instance, in the example introduced in section \ref{ME}, each \textit{Number of credits} is associated with the threshold value 0. When a partition on a \textit{Number of credits} variable happens, the partition containing the observations where the \textit{Number of credits} is strictly greater than 0 gain access to the corresponding \textit{Grade} variable as illustrated in figure \ref{GradeTree}.
 
\bigskip

Here is a pseudo-code of the proposed algorithm :
 
\begin{center}
\begin{tabular}{||l||}
\hline
\textbf{Algorithm} : BEST($S$,$\beta$,$V$) \\
\hline
\hline
0. Start with the entire data set $S$ and the set of available predictors $V[1]$.  \\
1. Check conditions (C1, C2 and C3). \\
2. \textbf{if}  (any condition is false) :  \\
\hspace{0.5cm} Assign a label to the node and exit.  \\
\textbf{ else} :  \\
\hspace{0.5cm} \textbf{for} ($j$ in all available predictors):  \\
\hspace{1.0cm} \textbf{for} ($s$ in all possible splits) : \\
\hspace{1.5cm} Compute total impurity measure. \\
\hspace{0.5cm} Select the variable $j$ and the split $s$ with minimum impurity measure. \\
\hspace{0.5cm} Split the node into two child nodes.\\
\hspace{0.5cm} Update the available predictors for both children nodes using $V[j+1]$. \\
\hspace{0.5cm} Repeat steps 1 \& 2 on the two children nodes.\\ 
\hline
\end{tabular}
\end{center}

\bigskip

In addition, the resulting tree can be pruned, and constructed with any splitting rule, any stopping rule and any label assignment rule. Since one of the goals of this new algorithm is to produce accurate but also interpretable models we did not discuss forests so far, but the proposed tree construction procedure can be used to build any type of forest as well. Our implementation is publicly available on CRAN under the package named \textit{BESTree} or on the first author's website. Anyone can download and install the package, read the vignette and use our proposed algorithm for their own research.    

\subsection{Theoretical justification}

Formally, the loss of a classifier $h$ is defined as :

\begin{align}
L_\mathcal{D}(h) = \mathbf{P}_{\mathcal{D}}[h(x_i) \neq y_i ],
\end{align}
which is the probability under the true data generating distribution $\mathcal{D}$ that the classifier $h$ misclassifies an observation $x_i$. Since the data generating distribution $\mathcal{D}$ is unknown, then the empirical loss computed with the data set $S$ is typically used as an estimator of the true loss :

\begin{align}
L_{S}(h) = \frac{|\{ i \in [n] : h(x_i) \neq y_i \}|}{n},
\end{align}
which is the proportion of misclassified observations in the training set $S$. Usually a set of classifiers $\mathcal{H}$ is selected in advance and most learning algorithms are trying to identify the classifier $h \in \mathcal{H}$ that minimizes the empirical loss $L_S(h)$. The set $\mathcal{H}$ was named the hypothesis class by Shai \& Shai \cite{Shai14}. The true loss can be decomposed in a manner to observe a bias-complexity tradeoff. Suppose $h_S = \underset{h \in \mathcal{H}}{\text{argmin}}L_S(h)$, then :

\begin{equation}
\begin{split}
L_D(h_S) &= \underset{h \in \mathcal{H}}{\text{min }}L_{\mathcal{D}}(h) + (L_D(h_S) - \underset{h \in \mathcal{H}}{\text{min }}L_{\mathcal{D}}(h)). \\
&= e_{\text{app}}(\mathcal{H}) + e_{\text{est}}(h_S).
\label{tradeoff} 
\end{split} 
\end{equation}

The approximation error, $\underset{h \in \mathcal{H}}{\text{min }}L_{\mathcal{D}}(h) = e_{\text{app}}$, is the minimum achievable loss within the hypothesis class.  The second term , $(L_D(h_S) - \underset{h \in \mathcal{H}}{\text{min }}L_{\mathcal{D}}(h))=  e_{\text{est}}$, is the estimation error and is caused by the use of the empirical loss instead of the true loss when selecting the best classifier $h$. Since the goal is to minimize the total loss a natural tradeoff emerges from equation \ref{tradeoff}. A vast, large and complex hypothesis class $\mathcal{H}$ leads to a wider choice of functions and therefore reduces $e_{\text{app}}$, but the classifier is more prone to overfitting, which increases $e_{\text{est}}$. Inversely, a small hypothesis class $\mathcal{H}$ reduces $e_{\text{est}}$ but increases $e_{\text{app}}$.

\bigskip

Our proposed algorithm aims at obtaining a better classifier by restricting the hypothesis class to a smaller one without increasing the approximation error. Suppose $\mathcal{H}_T$ is defined as the set of all tree-structured classifiers. Then, BEST is a new algorithm that aims to find the best classifier in a new hypothesis class $\mathcal{H}_{B}$ that contains all the tree-structured classifiers that respect a set of conditions regarding the order that variables can be partitioned upon. Therefore, we have $\mathcal{H}_{B} \subset \mathcal{H}_T$. Since the complexity of $\mathcal{H}_{B}$ is smaller than the complexity of $\mathcal{H}_T$ the estimation error of BEST will be smaller. 

\bigskip

Next, let us take a look at the approximation error  : $\underset{h \in \mathcal{H}_{B}}{\text{min }}L_{\mathcal{D}}(h)$. When using BEST, we make multiple assumptions on how the partitioning should be processed. For example, we assume it is better to partition the data using the missing indicator $M_j$ before partitioning the data using $X_j$. Doing so, we assume that the best tree-structured classifier among all classification tree $\mathcal{H}_T$ is contained within the set of tree-structured classifiers that respect the partition ordering that defines $\mathcal{H}_B$. In other words, we assume $\underset{h \in \mathcal{H}_T}{\text{argmin}}L_D(h) \in \mathcal{H}_{B}$. Suppose $S$ is a data set, $h_S(\mathcal{H}_T)$ is the classifier that minimizes the empirical loss on $\mathcal{H}_T$ and $h_S(\mathcal{H}_B)$ is the classifier that minimizes the empirical loss on $\mathcal{H}_B$, then :

\begin{equation}
\begin{split}
L_D(h_S(\mathcal{H}_T)) &= \underset{h \in \mathcal{H}_T}{\text{min }}L_{\mathcal{D}}(h)  + e_{\text{est}}(h_S(\mathcal{H}_T)). \\
&= \underset{h \in \mathcal{H}_B}{\text{min }}L_{\mathcal{D}}(h)  + e_{\text{est}}(h_S(\mathcal{H}_T)). \\
&\geq \underset{h \in \mathcal{H}_B}{\text{min }}L_{\mathcal{D}}(h)  + e_{\text{est}}(h_S(\mathcal{H}_B)). \\
&=  L_D(h_S(\mathcal{H}_B)),
\label{proof}
\end{split} 
\end{equation}

\setlength{\parindent}{0cm}

which implies that the under the assumption we have made we would not only naturally manage missing values but also reduce the loss. If our assumption $\underset{h \in \mathcal{H}_T}{\text{argmin}}L_D(h) \in \mathcal{H}_{B}$ is false, we might increase the loss, and the assumption itself is impossible to verify. Therefore, the behaviour of the algorithm under multiple scenarios will be tested in section \ref{Sec:Res} with simulated data.

\setlength{\parindent}{15pt}

\section{Related work} \label{litrev}

Decision trees are well-established and a wide variety of solutions has already been proposed to handle missing values. In this section, we will position our contribution within the current literature. We will briefly introduce the current missing value techniques that are paired with decision trees and we will establish the main differences between these techniques and the proposed algorithm introduced in section \ref{BEST}. To do so, we will use recent surveys \cite{Saar07,Twala09,Ding10,Gavankar15} that defined and compared these techniques using various simulated and real data sets.

\bigskip 

Predictive value imputation (PVI) methods are popular approaches to deal with missing values. They estimate and impute the missing values within both the training and the test set. The simplest imputation consists of replacing the missing values with the mean for numerical predictors or the mode for categorical predictors. More advanced prediction models have also been proposed, such as linear model, k-nearest neighbours or expectation-maximization (EM). 

\bigskip

Since these methods use known predictors to impute values for the missing ones, if the predictors are uncorrelated these approaches will have no predictive power. This will lead to poor imputation and it is a major drawback noted by Saar-Tsechansky and Provost \cite{Saar07} and Gavankar \cite{Gavankar15}. Nonetheless, Twala \cite{Twala09} demonstrated using simulated data sets the great performances of expectation-maximization multiple imputations (EMMI) \cite{Schafer97}.  This imputation technique produces multiple different imputations based on expectation-maximization and then aggregates the results. 

\bigskip

Our proposed algorithm differs from imputation methods as it only uses known information to build the classifier instead of using potentially unreliable prediction to replace missing values.

\bigskip

The surrogate variable (SV) approach \cite{Breiman84} is a special case of predictive value imputation. As explained by Hastie et al. \cite{Hastie09}, during the training process, when considering a predictor for a split, only the observations for which that predictor is not missing are used. After the primary predictor and split point have been selected, a list of surrogate predictors and split points is constructed. The first surrogate split is the predictor and split point pair that best mimic the split of the training data achieved by the primary split. Then the second surrogate split is determined among the leftovers predictors and so on. When splitting the training set during the tree-building procedure or when sending and observation down the tree during prediction, the surrogate splits are used in order if the primary splitting predictor value is missing. 

\bigskip

Many articles \cite{Feelders99,Saar07,Twala09,Ding10} showed that the results are not satisfactory in many cases and Kim and Loh \cite{Loh01} noted the variable selection biased caused by this approach. Our proposed approach is much more computationally efficient and utilizes the missing pattern as a predictor instead of ignoring it.

\bigskip

Reduced-feature models are suggested by Saar \cite{Saar07} when missing values appear only in the prediction process. When we need to classify a new observation, a tree is built using only the known predictors of the new observation. If multiple observations contain different missing pattern then multiple trees are built to classify the various observations. It shares a great deal of similarities with lazy decision trees \cite{Friedman97} as both models tailor a classifier to a specific observation and uses only known predictors to do so.

\bigskip

A major drawback of this technique is that it only manages missing values during prediction while our proposed technique can handle missing value for both training and prediction. BEST also differs from reduced-feature models as it not only uses the known values but also utilizes the fact that we know some predictors are missing instead of discarding this information. 

\bigskip

The popular C4.5 implementation \cite{Quinlan93} has its own way to manage missing data, defined as a distribution-based imputation (DBI). When selecting the predictor to split upon, only the observations with known values are considered. After choosing the best predictor to split upon, observations with known values are split as usual. Observations with unknown values are distributed among the two child nodes proportionately to the split on observed values. Similarly, for prediction, a new test observation with missing value is split intro branches according to the portions of training example falling into those branches. The prediction is then based upon a weighted predictions among possible leaves.

\bigskip

This technique is computationally slow and offers poor performances in terms of prediction accuracy according to some of the surveys we mentioned \cite{Saar07,Twala08}. Our technique should be faster, more accurate and more interpretable.

\bigskip

As described by Ding and Simonoff \cite{Ding10}, the Separate Class (SC) method replaces the missing value with a new value or a new class for all missing observations. For categorical predictors we can simply define \textit{missing value} as a category on its own and for continuous predictors any value out of the interval of observed value can be used. 

\bigskip

This technique has the best performances according to Ding and Simonoff \cite{Ding10} when there are missing values in both the training and the test set and when observations are missing not at random (MNAR). Twala et al. \cite{Twala08} also came up with similar results with a generalization of the separate class method coined Missing Incorporated in Attribute (MIA). These techniques are by far the closest in spirit to BEST. As BEST, MIA and SC allow for similar data partitioning we do not expect BEST to offer a drastic improvement in terms of accuracy. On the other hand, BEST identifies the missing pattern using other predictors rather than including the missingness information within the predictor containing missing values. Doing so, our approach should offer highly interpretable results and a more accurate variable importance analysis.

\bigskip

Finally, there exist many other articles discussing decision trees in the context of missing values. Some introduce ways to use decision trees and random forests as missing values imputation techniques such as Rahman and Islam \cite{Rahman13} and others to identify the missingness structure such as Tierney et al. \cite{Tierney15}. Our approach is different since we are not using decision trees to pre-process data with missing value or to identify the missingness structure but to perform a decision tree analysis of a data set that contains missing values.

\section{Experiments : Simulated data sets} \label{Sec:Res}

Now we are going to asses the abilities of our algorithm on simulated data sets. All of our experiments are done using the R-language \cite{R}. In the following set of simulations, we will compare 6 methods; (1) BEST, our proposed approach, (2) the Distribution Based Imputation (DBI) proposed by Quinlan \cite{Quinlan93} implemented in the \textit{C5.0} package \cite{Kuhn18}, (3) a simple single variable imputation (SVI), either the mode for a categorical predictor or the mean for a numerical one, (4) a refined predictive value imputation (PVI) using known predictors; EM for numerical predictors and multinomial logistic regression for categorical ones \cite{van11}, (5) the separate class (SC) approach and finally (6) the surrogate variable technique introduced by Breiman et al. \cite{Breiman84} implemented in the \textit{rpart} package \cite{Therneau18}. Since the Reduced-Feature Model was the least accurate in every single experiment we have done, we decided not to include it in the following plots to improve readability.

\bigskip

\begin{figure}
\begin{tikzpicture}[
    scale = 1, transform shape, thick,
    every node/.style = {draw, circle, minimum size = 8mm},
    grow = down,  
    level 1/.style = {sibling distance=6.6cm},
    level 2/.style = {sibling distance=3.3cm}, 
    level 3/.style = {sibling distance=1.4cm}, 
    level 4/.style = {sibling distance=2cm}, 
    level distance = 2.0cm
  ]
  \node (Start){} {
   child  { node (A) {}
     child { node (B) {}
          child { node  (G) {1}}
     	  child { node  (H) {2}}
     }
     child { node (C) {}
          child { node  (I) {3}}
          child { node  (J) {4}}
     }
   }
   child {   node  (D) {}
     child { node  (E) {}
          child { node  (K) {5}}
          child { node  (L) {6}}
     }
     child { node  (F) {}
          child { node  (M) {7}}
          child { node  (N) {8}}
     }
   }
  };

  \begin{scope}[nodes = {draw = none}]
    \path (Start) -- (A) node [near start, left]  {\footnotesize $X_1 = 0$};
    \path (A)     -- (B) node [near start, left]  {\footnotesize $X_2 = 0$};
    \path (A)     -- (C) node [near start, right] {\footnotesize $X_2 = 1$};
    \path (Start) -- (D) node [near start, right] {\footnotesize $X_1 = 1$};
    \path (D)     -- (E) node [near start, left]  {\footnotesize $X_4 \leq 0.3$};
    \path (D)     -- (F) node [near start, right] {\footnotesize $X_4 > 0.3$};
    \path (B)     -- (G) node [near start, left]  {\footnotesize $X_4 < 0.7$};
    \path (B)     -- (H) node [near start, right] {\footnotesize $X_4 \geq 0.7$};
    \path (C)     -- (I) node [near start, left]  {\footnotesize $X_3 < 0.6$};
    \path (C)     -- (J) node [near start, right] {\footnotesize $X_3 \geq 0.6$};
    \path (E)     -- (K) node [near start, left]  {\footnotesize $X_2 = 0$};
    \path (E)     -- (L) node [near start, right] {\footnotesize $X_2 = 1$};
    \path (F)     -- (M) node [near start, left]  {\footnotesize $X_3 \leq 0.2$};
    \path (F)     -- (N) node [near start, right] {\footnotesize $X_3 > 0.2$};
    
  \end{scope}
\end{tikzpicture}
\caption{The decision tree used to generate our simulation data sets\label{Gen}}
\end{figure}
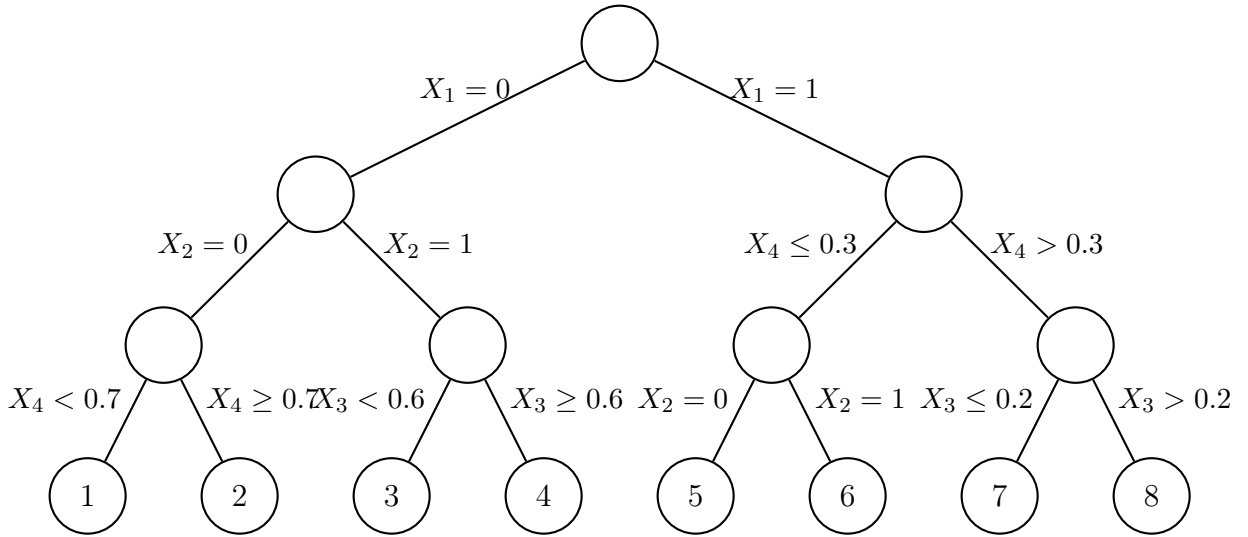

We will generate data sets containing 4 predictors; $X_1$ and $X_2$ are binary predictors and $X_3$ and $X_4$ are continuous predictors on $(0,1)$. The response is categorical and can take up to 8 different values. The binary predictors are generated according to a Bernoulli distribution and the continuous predictors are generated according to a Uniform distribution. The response labels are assigned according to the tree in figure \ref{Gen} but a proportion of the responses labels are randomly assigned. 

\bigskip

Let us describe the experimental procedure. We begin by generating a data set as we described in the previous paragraph. We will apply a missing pattern to the data set; details are included in the respective subsections. Then we will fit a pruned decision tree using each of the six methods mentioned earlier in the section.  We will adjust various parameters such as the number of training observations, the variable containing missing values and the type of missing patterns. We will compare the technique performances using  the prediction accuracy on the test set where we define the accuracy as the proportion of correctly classified observations.

\subsection{MAR : Missingness depends on observed predictors} 

This first experimental set up is meant to test the missing pattern structure BEST was designed for. In this set up, the missing pattern of a predictor is fully explained by another, fully observed, predictor. The binary predictor $X_1$ was designated as gating variable for a randomly sampled predictor, either $X_2$, $X_3$ or $X_4$.  In our first experiment we randomly sample either 0 or 1 and the gated variable is missing if $X_1$ equals the randomly sampled value. This procedure will be repeated 200 times for this experiment.
\bigskip

We will present our results using Sina Plots \cite{Sidiropoulos15,Wickham16}. This will allow to better visualize the distribution of the performances of the different techniques for different predictors with missing values.

\begin{figure}[H]
\centering
\includegraphics[height=12cm,width=14cm]{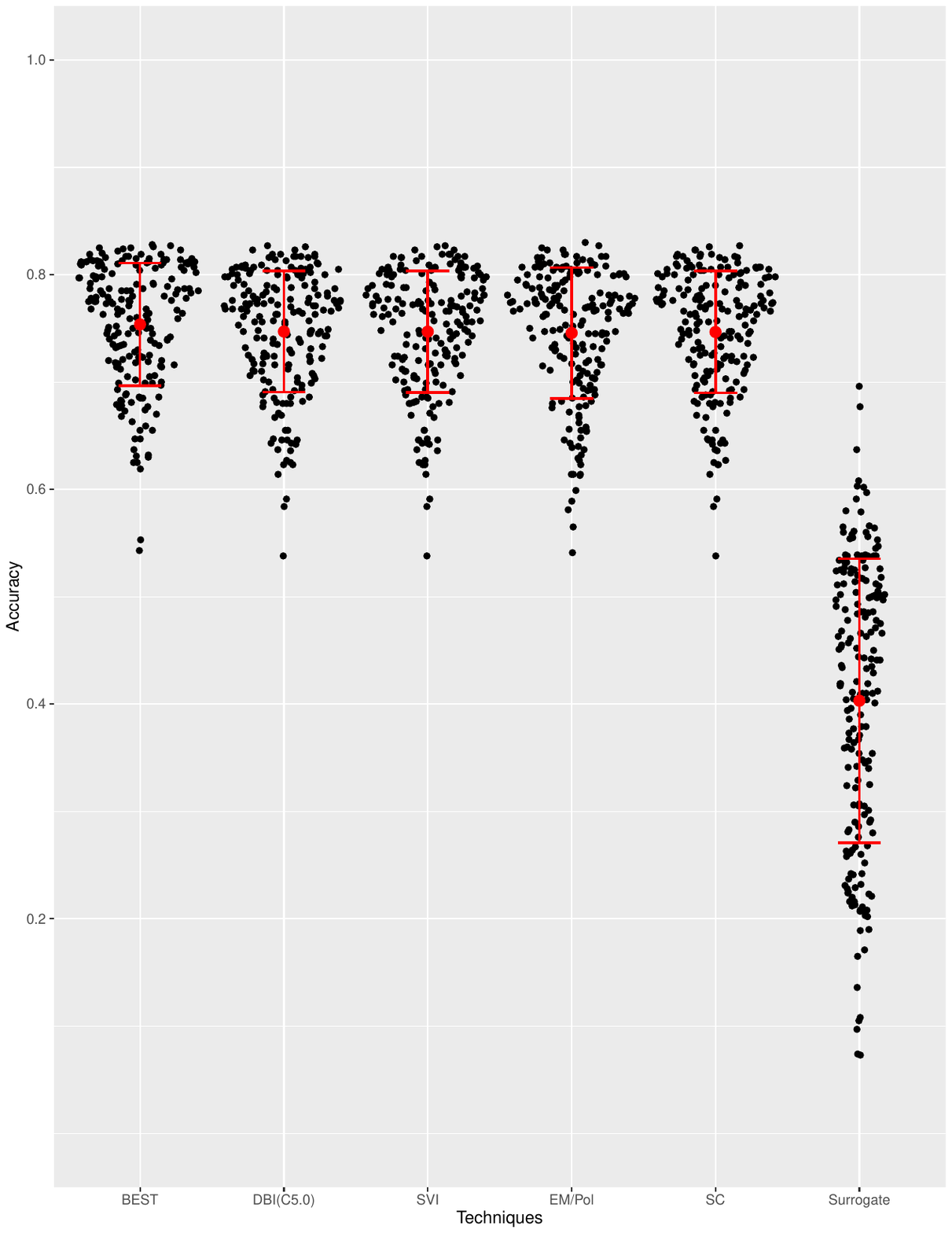}
\caption{Accuracy by techniques when the missing pattern depends on other predictors with 200 training observations  \label{MARX200} }
\end{figure}

\begin{figure}[H]
\centering
\includegraphics[height=12cm,width=14cm]{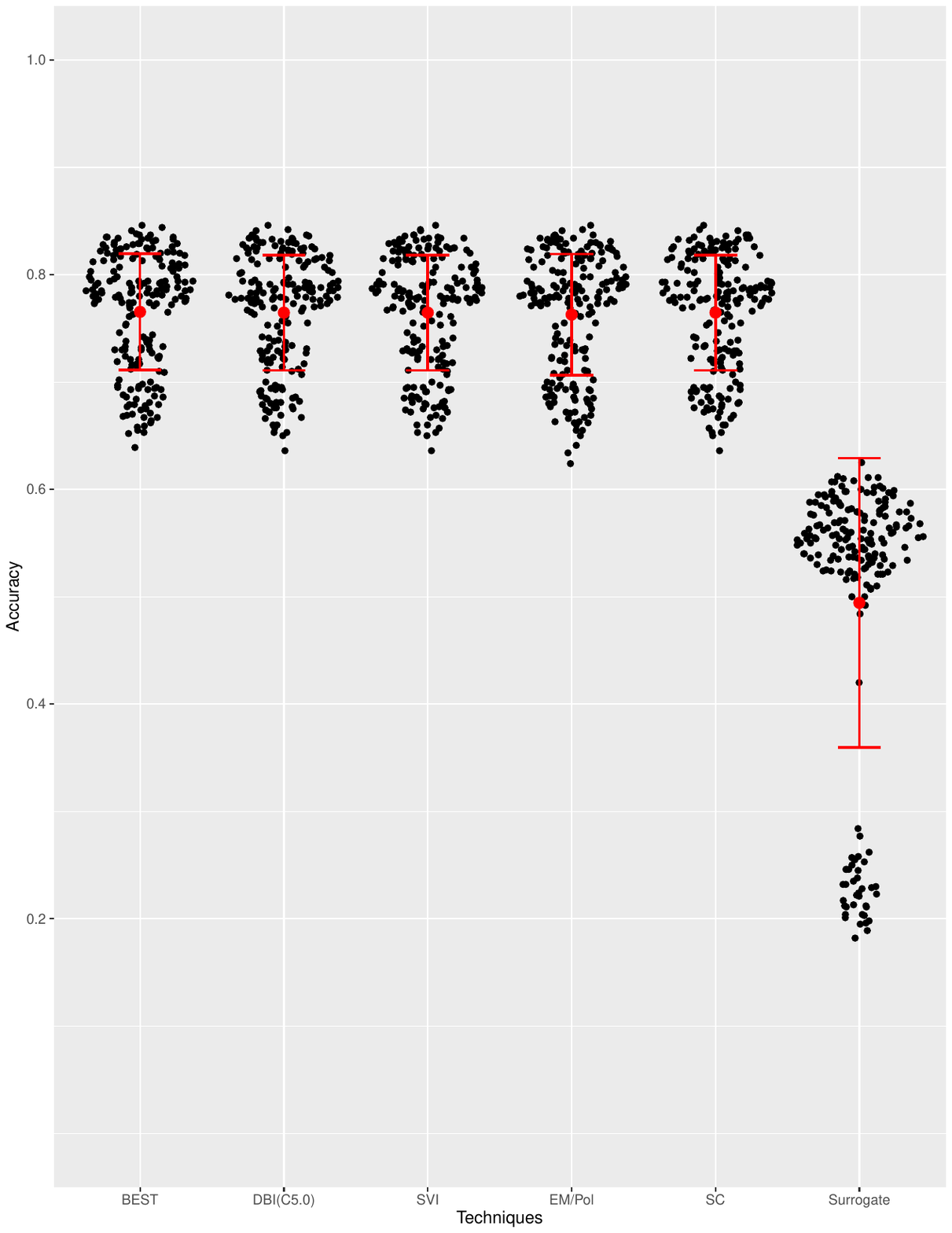}
\caption{Accuracy by techniques when the missing pattern depends on other predictors with 1000 training observations \label{MARX1000} } 
\end{figure}

The figure \ref{MARX200} contains the results of our simulation when we the techniques used 200 training observations to build the classifiers. Figure \ref{MARX1000} contains the results of our simulation obtained when we used 1000 training observations.

\bigskip

Both of the plots convey similar information. When the missing pattern depends on other predictors, the performance of BEST is similar to many competitors but BEST still leads to more interpretable decision trees and does not require any imputation. There is no notable differences between the results obtained with 200 observations and 1000 observations in that precise experiment. Going forward we will keep the number of training observations fixed to 200 but we will vary other parameters.

\bigskip

In the next experiment we will once again designate $X_1$ as the gating variable and will again sample one of the other predictors as the gated predictor. Once again, we randomly sample either 0 or 1 but this time, we will generate as less extreme missing pattern. The gated variable will be missing with a probability of 50\% when $X_1$ equals the randomly sampled value.

\begin{figure}[H]
\centering
\includegraphics[height=12cm,width=14cm]{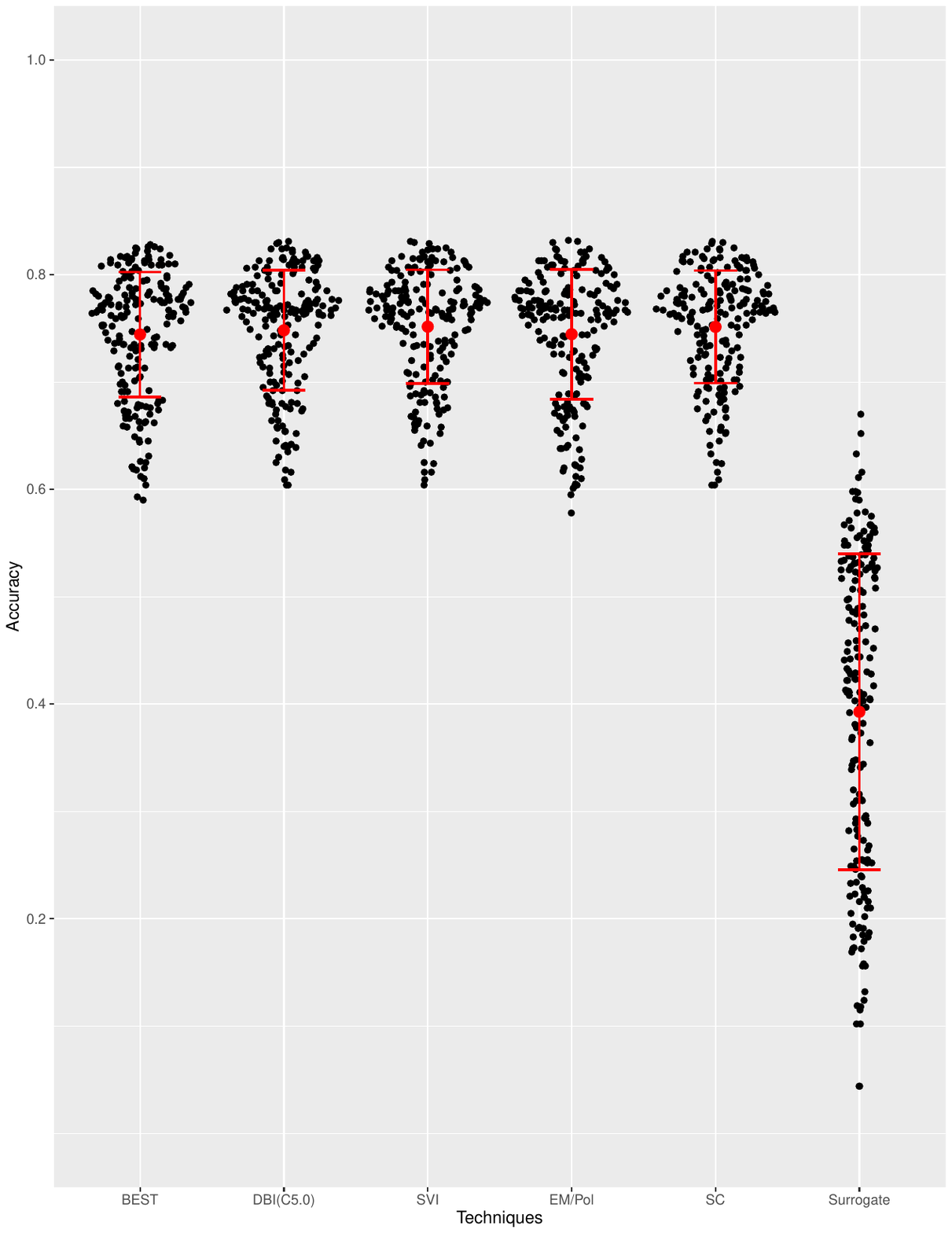}
\caption{Accuracy by techniques with 50\% of the gated variable missing when the condition of the gating variable is respected \label{MARX05} }
\end{figure}

We can observe in figure \ref{MARX05} that a less extreme missing pattern does not affect one technique more than others.

\bigskip

Overall, BEST performances when the data is missing at random given other predictors is as good as other tested techniques.

\subsection{MAR : Missingness depends on the response} 

According to Ding et al. \cite{Ding10}, the relationship between the missing pattern and the response variable has a great effect on the results obtained from different missing value treatments. 

\bigskip

 In this simulation, one of the predictors is randomly selected, let us say $X_j$, every iteration and the censoring process is then applied. The censoring process goes as follows; one of the eight response labels is randomly selected, and $X_j$ is missing for all observations with that selected label. In this experiment we have used a dummy variable as the gating variable for the BEST algorithm. This procedure will be repeated 100 times for this experiment.
 
\begin{figure}[H]
\centering
\includegraphics[height=12cm,width=14cm]{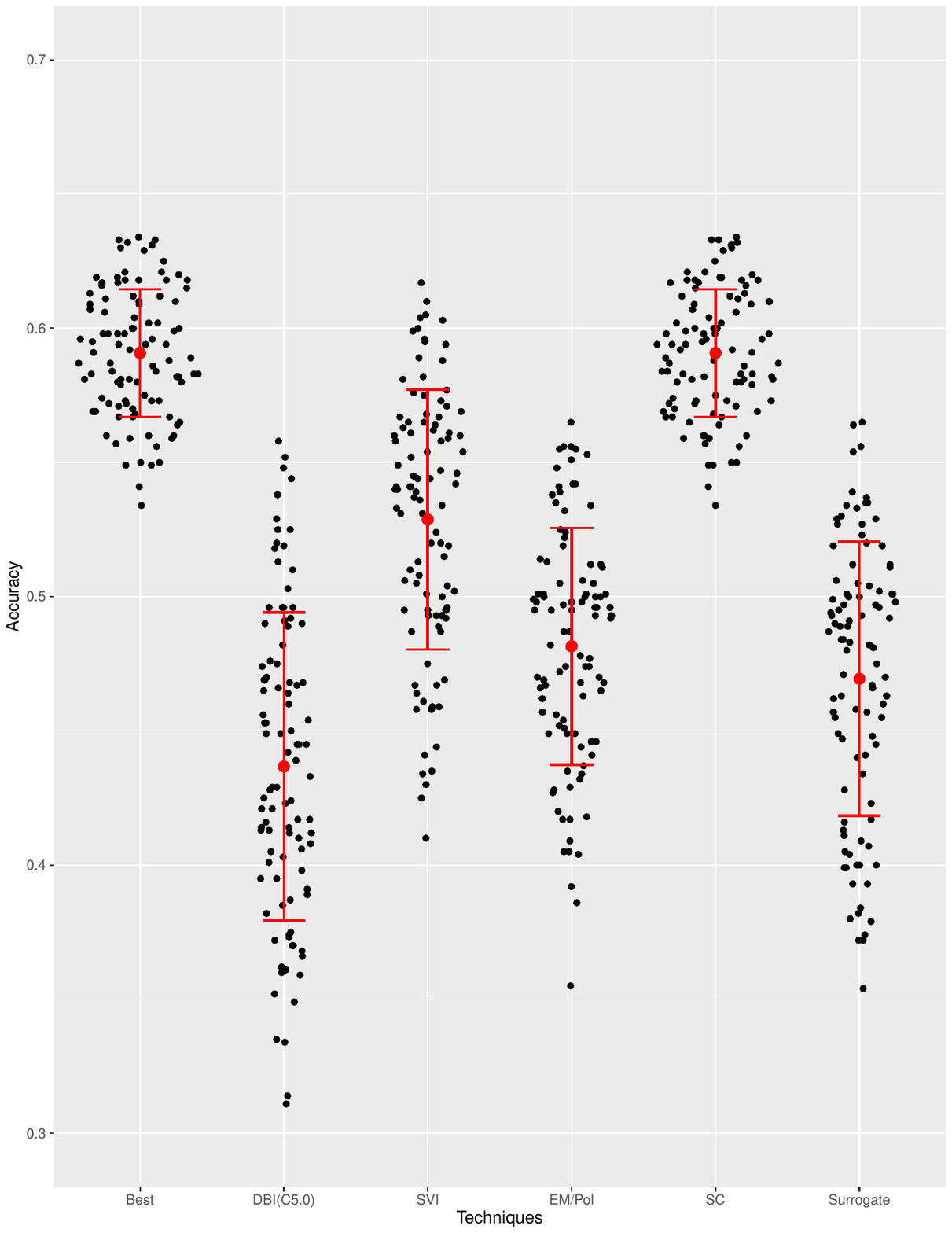}
\caption{Accuracy by techniques when the missing pattern depends on the response. \label{MARY} }
\end{figure}

In this experiment, the missing pattern is actually a variable with predictive power and therefore, models like BEST and SC shine as they utilize the fact that there is missing values instead of trying to impute them. BEST and SC approaches have similar results and their performances is higher than any other techniques. 

\bigskip

It is interesting to notice the high performances of the simple single value imputation in some cases. Our experiments revealed that when the predictor containing missing value is continuous, replacing missing values with the mean is equivalent to creating a separate class because only the missing values will exactly take the value of the mean. If the predictor with missing value is categorical, replacing missing values with the mode will make the observation with missing value undistinguishable from observations that truly belong to that class which leads to the poor results.

\bigskip

Next, we want to understand the impact of the indicator gating variable on the success of BEST. To do so, we added the indicator variable to the data set and we applied the various missing value techniques as well. 

\begin{figure}[H]
\centering
\includegraphics[height=12cm,width=14cm]{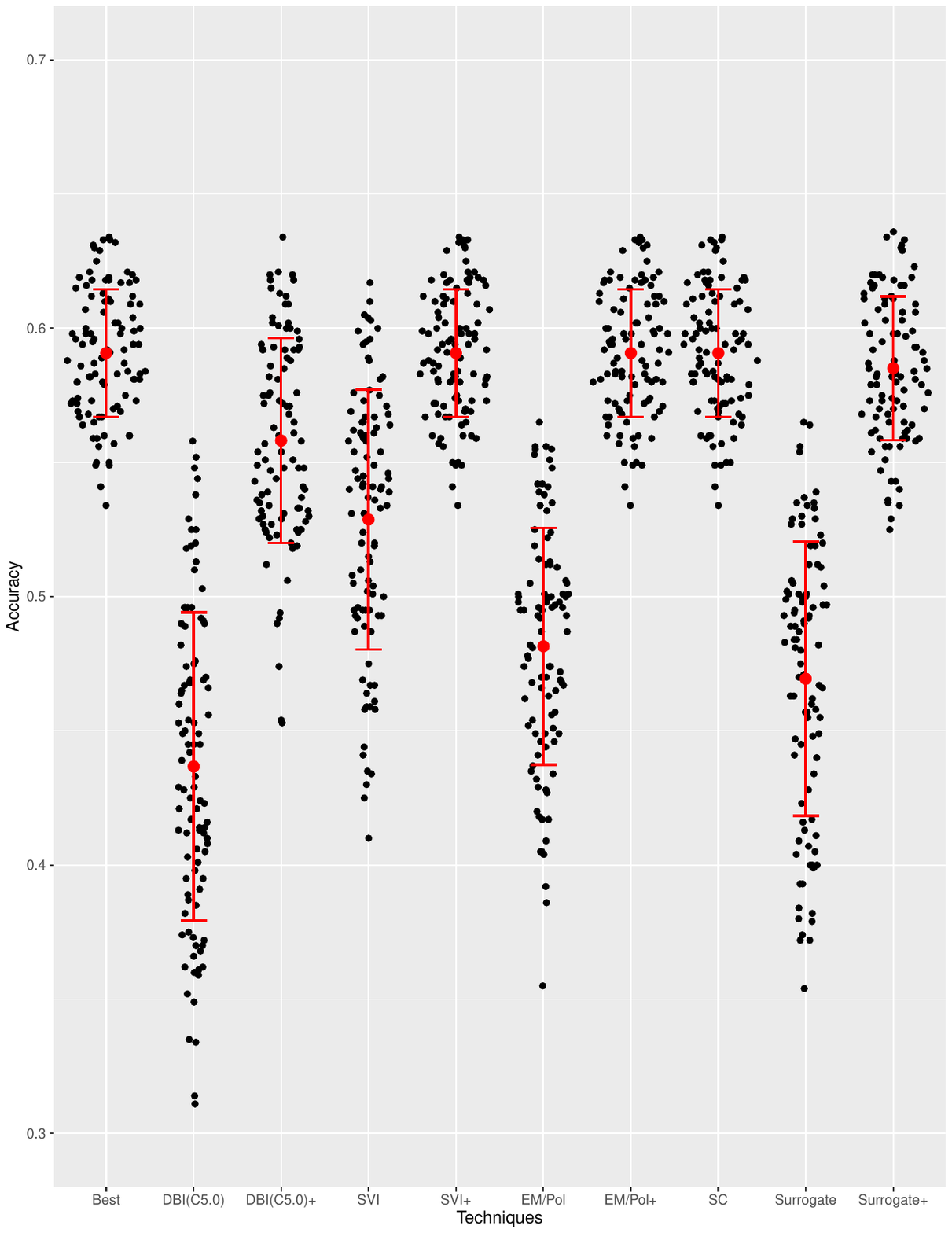}
\caption{Accuracy by techniques when the missing pattern depends on the response  \label{MARY_I} }
\end{figure}

In figure \ref{MARY_I} we can see that the indicator variable is important if the missingness depends on the response variable. In the figure above, the plus sign represents the result obtained by the technique with a data set that also includes the indicator variable. For example, for SVI+, we both imputed the missing values using SVI and we included the indicator variable in the data set. Adding this variable to the data set lead to improved performances for all of the tested techniques.  BEST offers great results considering it does not need to impute the missing values. If the missing indicator variable is to be added to the data set, it is counter-intuitive to also impute missing values. We argue in sections \ref{inter} and \ref{RFRes} that BEST also produces trees that are more interpretable and that BEST leads to a more reliable variable importance analysis than other algorithms. Thus, we believe BEST is the best approach when the pattern is missing at random given the response. 

\subsection{MNAR : Missingness depends on missing values}

Let us now proceed with an experiment when the predictors are missing not at random. If a continuous predictor, let us say $X_j$, is randomly selected, then a random value $t$ which serves as threshold within the domain of $X_j$ is also drawn at random. Finally, a Bernoulli variable $b$ is drawn and if $b=0$, then if $X_j < t$ $X_j$ is set missing, otherwise if $b=1$ then $X_j$ is missing if its value is greater than $t$.

\bigskip

If one of the binary predictors is selected, then a Bernoulli variable is drawn and $X_j$ is missing if $X_j$ equals the Bernoulli variable. Since the missingness of $X_j$ depends on the value of $X_j$ itself, then this is MNAR. The procedure was repeated 100 times. 

\begin{figure}[H]
\centering
\includegraphics[height=12cm,width=14cm]{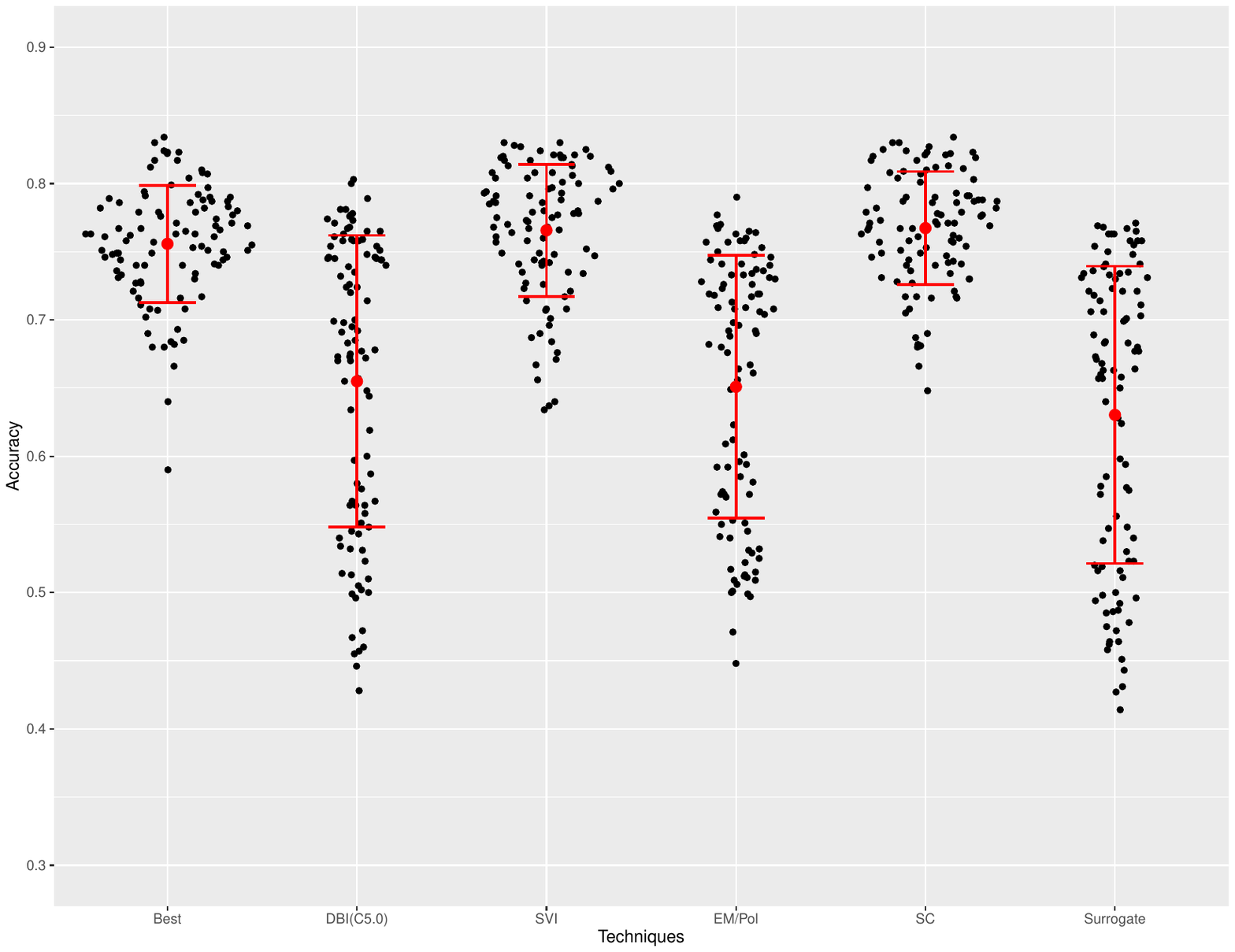}
\caption{Accuracy by techniques when the missing pattern depends on missing values  \label{MNAR} }
\end{figure}

In figure \ref{MNAR} we observe that BEST outperforms DBI and multiple imputations. The performances of BEST are comparable to those of SVI but are slightly under those of the SC approach.

\subsection{Random forests and variable importance} \label{RFRes}

Let us now build a small example where random forests are used to analyse the variable importance. Random forests are popular in exploratory analysis \cite{Strobl07} as the variable importance tools that were developed for this model became popular. 

\bigskip

As we have seen in the previous experiments, when BEST performs well, so does the SC approach usually as both of these techniques produce similar trees. Here we will quickly discuss how BEST produces more accurate variable importance computations than the SC approach. We have created an example where the missing pattern depends on the response, used either the SC approach or BEST to handle missing values and we have built a forest of those trees. 

\bigskip

When the values for a predictor are conditionally missing at random given the response, the missing pattern is itself a good predictor. We believe it is important that a variable importance analysis distinguishes between the importance of the predictor with missing value, say $X_j$, from the importance of its missing pattern $M_j$. A random forest of trees built under the SC approach would fail to distinguish between the effect of the observed value for that predictor and the effect of the missing pattern. Since BEST actually uses a variable to define the region with missing values, either with another predictor or a user-created dummy variable, then this gating variable importance will better represent the predictive power of the missing pattern. 

\begin{table}[H] 
\begin{center}
\resizebox{\textwidth}{!}{
\begin{tabular}{ |c|c|c|c|c|c|c|c|c|c| }  
\hline
 Data & $X_1$ & $\mathbf{X_2}$ & $X_3$ & $X_4$ & $\mathbf{M_2}$ \\
\hline
Complete & 112.37705 & \textbf{26.76542} & 158.55069 & 102.91328 & - \\ 
BEST &  94.53228 & \textbf{10.80073} & 155.39201 & 78.14924 & \textbf{156.27593} \\
SC & 103.62706 & \textbf{171.70715} & 147.89474  & 80.51999 & -  \\ 
SC+. & 94.36789 & \textbf{81.83499}  & 128.78712 &  55.14764 &  \textbf{84.45193}  \\ 
\hline
\end{tabular}}
\end{center}
\caption{Variable Importance table computed using the GINI decrease importance}
\label{VITab} 
\end{table}

We have built a random forest using the complete data set and computed the GINI decrease importance. Then we have randomly selected one of the eight labels, and the predictor $X_2$, a predictor of low importance according to the GINI decreases under the complete data set, is rendered missing depending on the value of the response. Since the SC approach uses the predictor containing missing value to identify observations containing missing value then it identifies $X_2$ as the most important predictor. Even when we include the missing pattern $M_2$ to the data set the SC approach uses, the variable importance for $X_2$ is still inflated. 

\bigskip

Using BEST, we can easily observe that the missing pattern, $M_2$ is the important predictor and that $X_2$ is actually of low importance when observed as it should be according to the complete data variable importance. We believe the variable importance analysis produce by BEST reflects better the true predictive power of the predictors and that this is a great benefit from using BEST over the SC approach.

\subsection{Simulations: takeaways and limitations}

Throughout various simulation experiments we have been able to highlight the success of BEST under various scenarios. Regarding the SC approach, since it does not impute the missing values then it can create the same partitioning that BEST creates but the SC approach does not need to first isolate the missing value in order to partition upon the variable containing missing values which can sometimes be valuable.

\bigskip

Our simulation revealed that BEST suffers from a weakness when the gating variable is of low importance. This can happen if only a small proportion of data is missing, if the missing pattern is simply non-informative or in some cases when data is MCAR. In that case, BEST will never partition upon the gating variable and thus will never partition upon the branch-exclusive variable which will almost surely reduce the accuracy of the resulting tree. This weakness is intrinsic to the algorithm as it is caused by the greedy nature of decision trees which are usually fitted by growing large trees and pruning them later. During the partitioning process, a classic decision tree approach only sees the reduction in impurity gained with a single partition and thus cannot perceive the accuracy gained by the combination of two successive partitioning. 

\bigskip

When facing this problem, using a random forest of BEST trees is a possible solution. When we build trees in a forest, predictors are randomly selected and thus the algorithm will partition on the gating variable of low important from time to time which will reveal the important gated variable. One way to entirely negate this limitation would be to consider pairs of consecutive splits but this would come at a great cost, it would drastically increase the run time of the algorithm. 

\bigskip

A limitation of our simulated experiments is the absence of the MIA algorithm \cite{Twala08} who shares a lot of similarities with BEST. No implementation or package of MIA was available at the time we ran this experiment and thus an advantage of BEST is that a R package is available for researcher looking for an algorithm ready to use off-the-shelf. Regarding prediction accuracy, we do not expect BEST to outperform MIA as they can produce similar partitioning but we expect BEST to be slightly faster as MIA greatly increases the number of operations when building the tree. Finally, as discuss earlier we do believe BEST produces more interpretable trees than the SC approach, the same argument holds when comparing BEST to MIA. The interpretability argument is expanded in section \ref{inter}.

\bigskip

Finally, another limitation of our experiment is that we did not test the run times of the algorithms in an extensive manner. We think there was no way to build a fair comparison among the various techniques. As it stands our package BESTree is entirely coded in R and thus currently runs more slowly than than popular decision tree packages such as the \textit{C5.0} package \cite{Kuhn18}, the \textit{rpart} package \cite{Therneau18}  or the \textit{party} package \cite{Hothorn06}. 

\bigskip

We believe that the speed difference is caused by our suboptimal coding and the language used but not by the structure of our proposed algorithm. Our BESTree package contains a regular decision tree function which shares most of its architecture with the BEST function but without the new functionalities introduced by BEST. We tested the run speed of our proposed algorithm BEST against the regular decision algorithm included in our package.  

\begin{figure}[H]
\centering
\includegraphics[height=8cm,width=12cm]{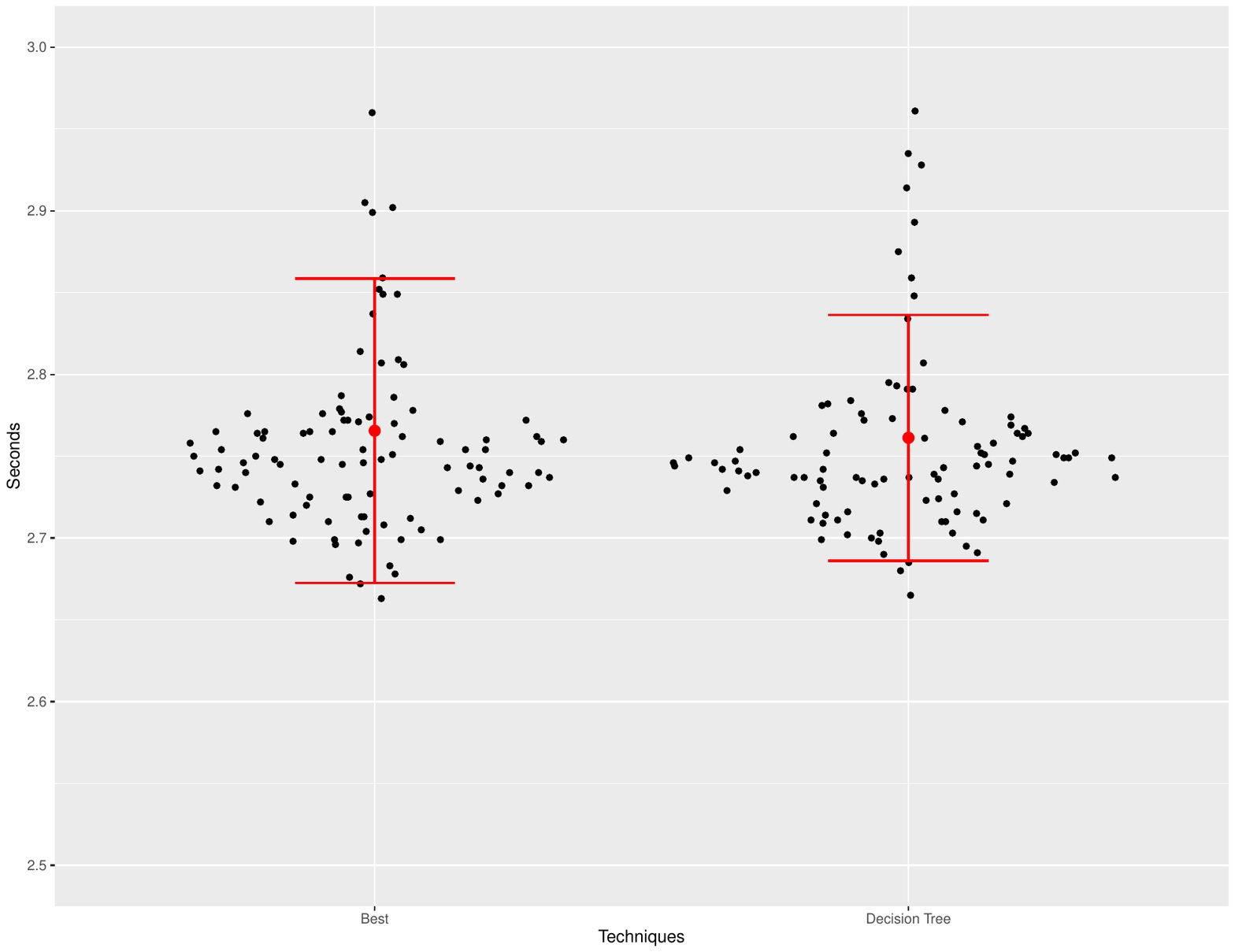}
\caption{Run time of BEST compared to a decision tree. \label{rt}}
\end{figure}

We generated 200 samples of 5000 observations and compared the run speed in seconds of BEST with a regular decision tree algorithm. As observed in figure \ref{rt}, our intuition was right; it seems like the added functionality of BEST does not increase the run time significantly. In order to improve the run time of all functions in our package, we will code some key components in a faster language such as C++ in a future release. 

\section{Experiments : grades data set}

\subsection{Predicting program completion}

The data set mentioned in section \ref{ME} was analysed using BEST. Once again, the accuracy of the proposed algorithm is compared to other techniques that handle missing values. To begin, we  predict if a student completes its program using its first year of courses and results. The data set contains 38842 observations. Our set of predictors consists of the number of credits attempted in all the departments and the average grade obtained in those respective departments. The number of credits is a numerical variable that serves as the gating variable for the respective grade. If the number of credits attempted in a department is greater than 0 for every observation in a region then BEST acquires access to the grade variable. We have randomly sampled training sets of different sizes and used all the remaining observations to assess the accuracy. This process was repeated 15 times and we have averaged the results. We did not include the single value imputation because we expect this technique to produce the same result as the SC approach since all predictors are numerical. We did not include the imputation produced by the mice package \cite{van11} as the package was incapable of handling the data set. Finally, we have used tables to show our results because with only 15 trials the Sina plots were not informative.

\begin{table}[H] 
\begin{center}
\resizebox{\textwidth}{!}{
\begin{tabular}{ |c|c|c|c|c|c|c|c|c|c| } 
\hline
 \ \# of obs   & \multicolumn{2}{|c|}{5000}  &\multicolumn{2}{|c|}{10000}  & \multicolumn{2}{|c|}{15000}  & \multicolumn{2}{|c|}{20000} \\
\hline
 \ Methods & Mean & S.D. & Mean & S.D. & Mean & S.D. & Mean & S.D.  \\
\hline
DBI & 0.7223 & 0.0035 & 0.7336 & 0.0038 & 0.7385 & 0.0092 & 0.7402 & 0.0095  \\
SC & 0.7307 & 0.0066 & 0.7387 & 0.0044 & 0.7427 & 0.0036 & 0.7462 & 0.0033  \\
Surrogate & 0.7291 & 0.0053 & 0.7301 & 0.0046 & 0.7311 & 0.0034 & 0.7300 & 0.0032 \\ 
BEST & 0.7333 & 0.0062 & 0.7424 & 0.0045 & 0.7457 & 0.0037 & 0.7479 & 0.0033   \\
\hline
\end{tabular}}
\end{center}
\caption{Mean accuracy and standard deviation when predicting program completion}
\label{Grade1} 
\end{table}

In table \ref{Grade1}, we observe that BEST produces the most accurate decision trees for that data set for all training data sizes. This improvement is important as it is essential for universities to predict if a student is at risk of not completing its program. This information is valuable since most universities want to help their students move forward by adjusting their policies or providing them the resources needed. We have done most of our experiments with trees, but using a random forest of BEST trees the prediction accuracy reaches 79.89\%, which is higher than anything previously obtained with competing techniques \cite{Beaulac18}.

\bigskip

Another reason why we might prefer using BEST in this analysis is its ability to rightfully identify the variable importance. As we discussed in section \ref{ME}, the researchers were interested in the importance of the predictors. Therefore BEST is an improvement as it truly identifies the importance of the gating variables, the number of credits, as we have shown in section \ref{RFRes}. In this case we were able to distinguish the importance of the number of credits in a department of the importance of the grade obtained in that department which was impossible before the implementation of BEST. In the variable importance plot below, the variable representing the number of credits in a department is identified by the department code, i.e. the number of credits in Chemistry is identified by CHM. The variable representing the averaged grade in a department is identified by the department code followed by the letter G, i.e CHM G represents the averaged grade in Chemistry.

\begin{figure}[H]
\centering
\includegraphics[height=9cm,width=\textwidth]{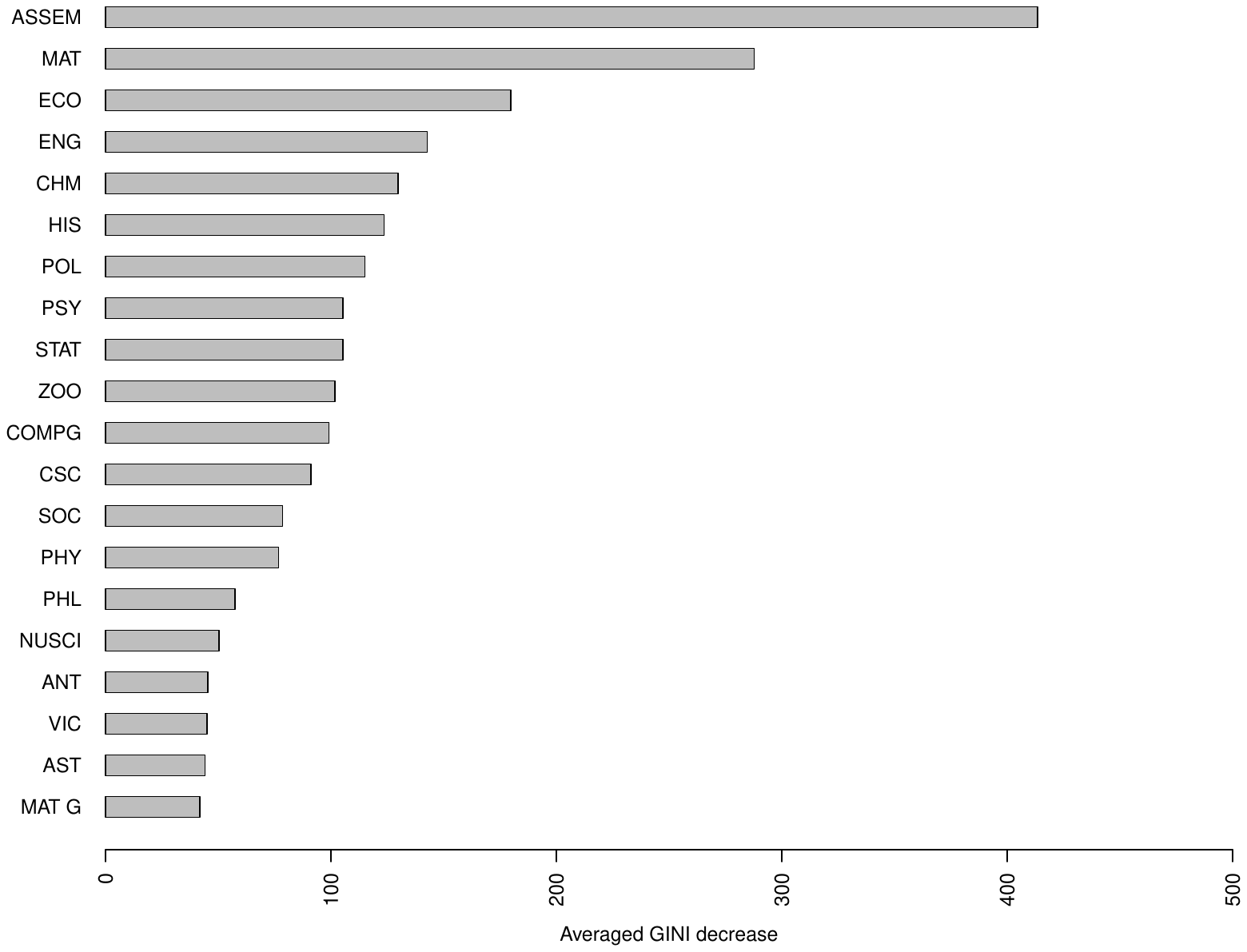}
\caption{Variable importance for the prediction of program completion. \label{COST_VIP_FoN} }
\end{figure}

Figure \ref{COST_VIP_FoN} illustrates the results obtained when we evaluated the variable importance using the averaged GINI decrease based on a forest of 200 BEST trees. The number of credits attempted in the first year seminar courses (ASSEM) is the most important predictor. In this case, taking a first-year seminar course was positively correlated with succeeding an undergraduate program. These seminars were brand new at the University at the time the data was collected and this analysis provides evidence of the merit of these courses to establish a student's profile.

\subsection{Predicting the major}

In the second analysis, we will look at the 26488 students who completed their program. Using the same set of explanatory variables, we will try to predict the department they majored in. Prediction the major that will be completed by students can help Universities plan ahead the resources needed by each departments. According the results in table \ref{Grade2} BEST can be helpful to produce such predictions:

\begin{table}[H] 
\begin{center}
\resizebox{\textwidth}{!}{
\begin{tabular}{ |c|c|c|c|c|c|c|c|c|c| } 
\hline
 \ \# of obs   & \multicolumn{2}{|c|}{5000}  &\multicolumn{2}{|c|}{10000}  & \multicolumn{2}{|c|}{15000}  & \multicolumn{2}{|c|}{20000} \\
\hline
 \ Methods & Mean & S.D. & Mean & S.D. & Mean & S.D. & Mean & S.D.  \\
\hline
DBI & 0.3581 & 0.0055 & 0.376 & 0.0043 & 0.3871 & 0.0039 & 0.3936 & 0.0057 \\
SC & 0.3992 & 0.0079 & 0.4172 & 0.0069 & 0.4262 & 0.0047 & 0.4332 & 0.0044 \\
Surrogate & 0.3639 & 0.0121 & 0.3661 & 0.011 & 0.3723 & 0.012 & 0.3742 & 0.012 \\ 
BEST & 0.3952 & 0.0074 & 0.4164 & 0.0051 & 0.4265 & 0.0043 & 0.4319 & 0.0043  \\
\hline
\end{tabular}}
\end{center}
\caption{Mean accuracy and standard deviation when predicting the major}
\label{Grade2} 
\end{table}

In table \ref{Grade2} we observe closer results from the two best-performing algorithms. BEST and SC have almost indistinguishable performances and are the top performers. A random forest of BEST trees reached an accuracy of 47.57\% which is slightly higher than previously obtained \cite{Beaulac18}. Even though the results are a lot closer between BEST and SC, our proposed algorithm still produces trees that are more interpretable and can be used to produce a non-biased variable importance analysis as argued in section \ref{RFRes}.

\begin{figure}[H]
\centering
\includegraphics[height=8cm,width=\textwidth]{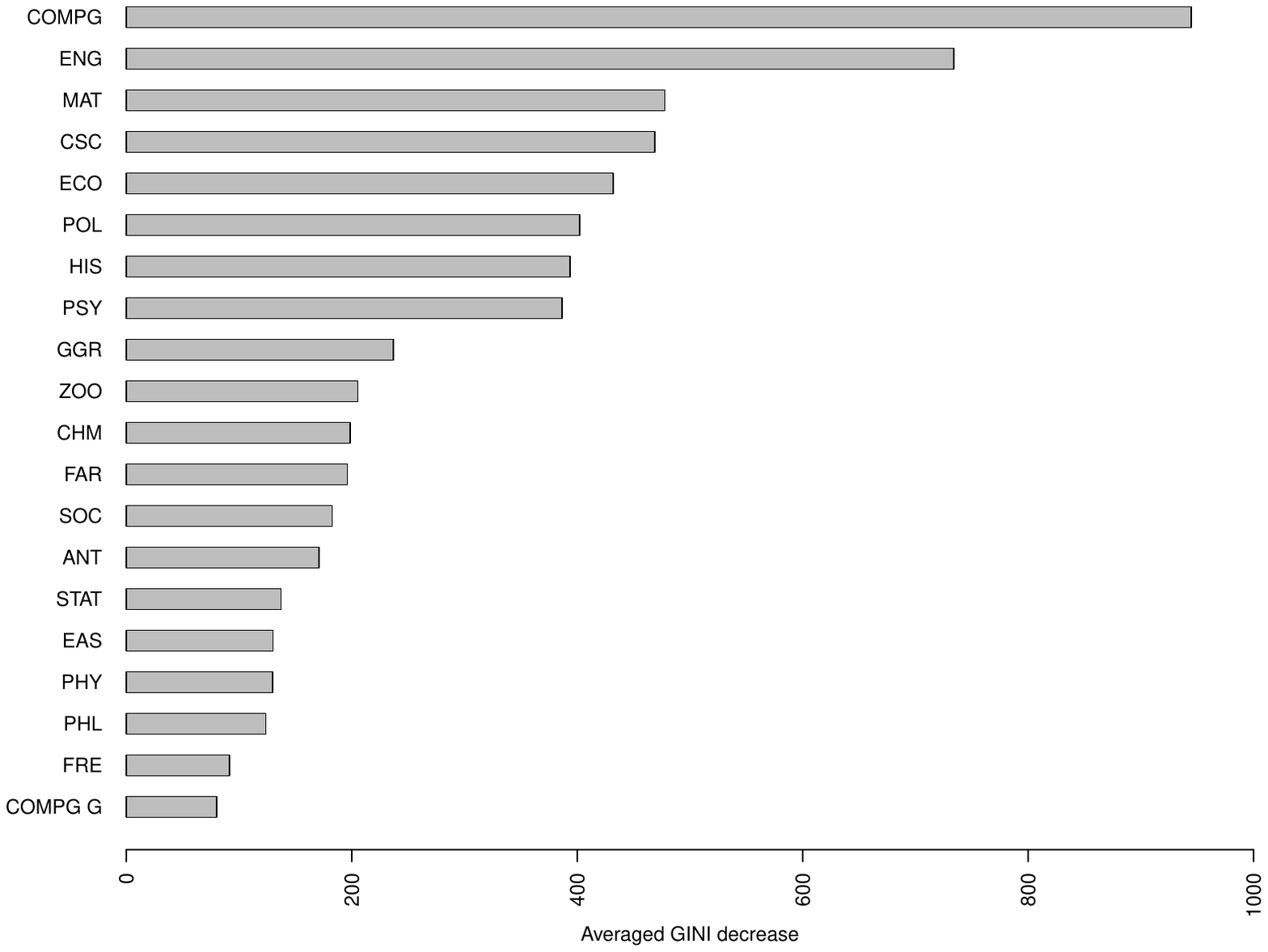}
\caption{Variable importance for the prediction of the major completed. \label{COST_VIP_Major} }
\end{figure}

Figure \ref{COST_VIP_Major} contains the variable importance we obtained. In this particular case the results were not significantly different from those previously obtained \cite{Beaulac18}. The number of credits in the Finance department (COMPG) and the number of credits in the English department (ENG) have relatively high importance compared to all the other predictors. We observed that the students who obtained a major in either of those were very likely to register to many courses in these respective departments starting in the first year. The number of credits in the Mathematics department (MAT) and the number of credits in the Computer Science department (CSC) are also of noticeably high importance. We have noticed that these variables are quite useful to predict if a student picks a major in a scientific field.  

\subsection{Improved interpretability} \label{inter}

We have mentioned throughout this article that we believe BEST leads to more interpretable decision trees than the SC approach. The experiments performed above provides a good example for that as they lead to interpretable trees.

\bigskip

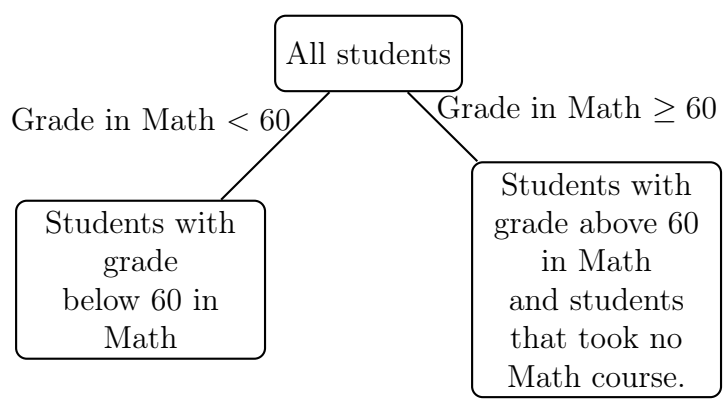
\begin{figure}[h]
\centering
\begin{tikzpicture}[
    scale = 1, transform shape, thick,
    every node/.style = {draw, rectangle, rounded corners, minimum size = 10mm},
    grow = down,  
    level 1/.style = {sibling distance=6cm},
    level 2/.style = {sibling distance=4cm}, 
    level 3/.style = {sibling distance=2cm}, 
    level 4/.style = {sibling distance=2cm}, 
    level distance = 2.5cm
  ]
  \node (Start) {All students} {
   child  { node (A) {\parbox{4cm}{\centering Students with grade below 60 in Math} } }
   child {   node  (D) {\parbox{4cm}{\centering Students with grade above 60 in Math and students that took no Math course.} }   }
  };

  \begin{scope}[nodes = {draw = none}]
    \path (Start) -- (A) node [near start, left]  { Grade in Math $< 60$};
    \path (Start) -- (D) node [near start, right] { Grade in Math $\geq 60$};

  \end{scope}
\end{tikzpicture}
\caption{An example of a decision tree partitioning produced by the SC approach and the associated regions \label{InterNo}}
\end{figure}

For the SC approach we have replaced missing grades by a value outside of the domain, 101. Frequently in this experiment, the tree constructed under the SC approach partitions upon the grade in departments before the number of credits. For example, if \textit{Grades in Mathematics} is the first split variable selected and 60 is selected as the split point, then the SC approach produces the partitioning illustrated in figure \ref{InterNo}.

\bigskip

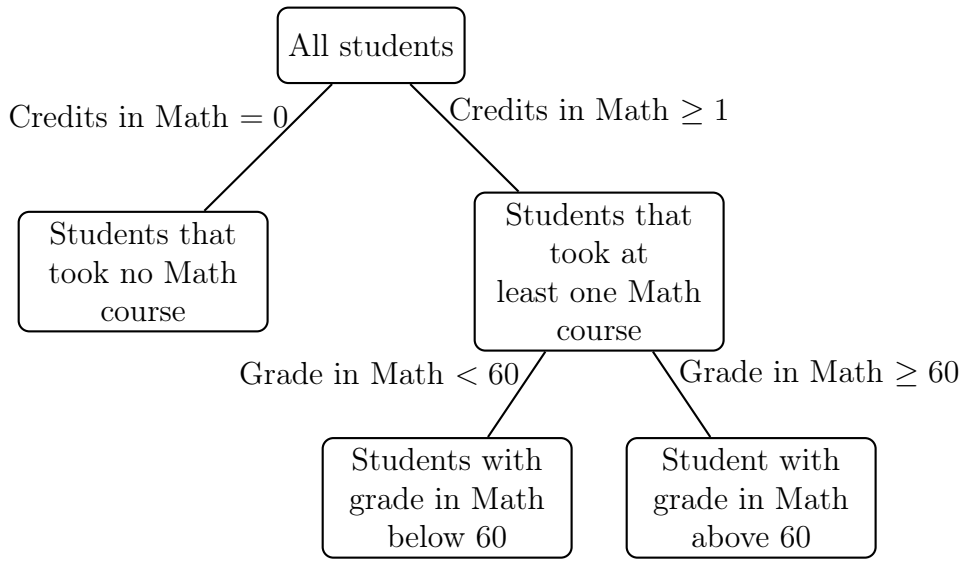
\begin{figure}[h]
\centering
\begin{tikzpicture}[
    scale = 1, transform shape, thick,
    every node/.style = {draw, rectangle, rounded corners, minimum size = 10mm},
    grow = down,  
    level 1/.style = {sibling distance=6cm},
    level 2/.style = {sibling distance=4cm}, 
    level 3/.style = {sibling distance=2cm}, 
    level 4/.style = {sibling distance=2cm}, 
    level distance = 2.5cm
  ]
  \node (Start) {All students} {
   child  { node (A) {\parbox{3cm}{\centering Students that took no Math course} } }
   child {   node  (B) {\parbox{3cm}{\centering Students that took at least one Math course} } 
   child { node (C) {\parbox{3cm}{\centering Students with grade in Math below $60$} }}
   child { node (D) {\parbox{3cm}{\centering Student with grade in Math above 60} }} }
  };

  \begin{scope}[nodes = {draw = none}]
    \path (Start) -- (A) node [near start, left]  { Credits in Math $= 0$};
    \path (Start) -- (B) node [near start, right] { Credits in Math $\geq 1$};
    \path (B) -- (C) node [near start, left] { Grade in Math $< 60$};
    \path (B) -- (D) node [near start, right] { Grade in Math $\geq 60$};

  \end{scope}
\end{tikzpicture}
\caption{An illustration of a decision tree partitioning produced by BEST and the associated regions \label{InterYes}}
\end{figure}

It is definitely hard to extract interpretable information out of this tree. Does this partition imply students with no experience in Mathematics behave similarly to students with good results in Mathematics ? BEST achieves similar or higher accuracy while keeping the partitions logical and interpretable. BEST will begin by partitioning students who attempted at least 1 credit in Mathematics from those who did not. Then, among students who attempted at least 1 credit in Mathematics, BEST will partition them according to their grades, which leads to a more interpretable sequence of partitions as illustrated in figure \ref{InterYes}.

\bigskip

If interpretability is considered a strength of decision trees, then BEST is better than the SC approach at preserving this strength. 

\subsection{Real-world data set experiment takeaways}

Even though we have performed experiments on a single real-world data, the results are extremely positive. BEST has higher or similar performance than the other tested algorithms. BEST produces more accurate variable importance analysis and more interpretable trees than the SC approach, BEST's closest competitor. Finally, to use BEST we did not need to do any imputations which is another reason why we prefer BEST.

\section{Conclusion}

We have constructed a modified tree-building algorithm that lets the users decide the regions of the predictor space where variables are available for the data partitioning process. We have focused on using this feature to manage missing values. BEST has the elegant property of analysing a variable only when values are known without assuming any missingness structure. It produces highly interpretable trees and achieves comparable accuracy to most missing value handling techniques in cases we have identified using simulated data sets. Even though BEST shares similarities with the separate class technique, BEST leads to a more accurate variable importance analysis and produces more interpretable and intuitive trees.

\bigskip 

BEST suffers from a weakness when the gating variable has no predictive power. In those cases, the algorithm will never choose to split upon the gating variable and thus will never be allowed to use the branch-exclusive variable. This problem can lead to a decrease in accuracy in some simple cases where the data is MCAR. Fortunately, as we have previously discussed, there already exist multiple techniques to handle data MCAR and we can count on cross-validation in order to help us select the best missing data handling technique. Nonetheless, in the simulated experiments we have performed, results were mostly positive as BEST outperforms some other techniques when data is MAR and MNAR.

\bigskip

The results produced by BEST were also satisfactory when the algorithm was used on the real motivating \textit{grades data set}. We were able to achieve higher accuracy than with most other techniques while obtaining a more interpretable classifier. Since variable importance was a concern in the \textit{grades data set} analysis, BEST was an improvement as it answers that research question by providing a more reliable variable importance analysis than the separate class approach previously used \cite{Beaulac18}. 

\pagebreak

\section*{Acknowledgement}

We are very grateful to Glenn Loney and Sinisa Markovic of the University of Toronto for providing us with the anonymised students grade data. The authors also gratefully acknowledge the financial support from NSERC of Canada. 

\bigskip

This is a pre-print of an article published in Computational Statistics. The final authenticated version is available online at: https://doi.org/10.1007/s00180-020-00987-z

\bibliographystyle{spmpsci}
\bibliography{mybibfile}


\end{document}